\documentclass[runningheads]{llncs}

\usepackage{eccv}

\usepackage{eccvabbrv}

\usepackage{graphicx}

\usepackage{booktabs}
\usepackage{microtype}
\usepackage{graphicx}
\usepackage{url}
\usepackage{xcolor}

\usepackage{graphicx}
\usepackage{wrapfig}
\usepackage{enumitem}
\usepackage{amsmath}
\usepackage{amssymb}

\usepackage{mathtools}
\usepackage{mathrsfs}
\usepackage{amsfonts}       
\usepackage{nicefrac}       
\usepackage{microtype}    
\usepackage{wrapfig}
         
\usepackage{booktabs}      

\usepackage{multirow}
\usepackage{caption}
\usepackage{tablefootnote}

\definecolor{mygray}{gray}{.9}

\usepackage{color, colortbl}

\usepackage{mwe}

\usepackage[accsupp]{axessibility}  

\usepackage{hyperref}

\usepackage{orcidlink}

\begin{document}

\title{How to Train the Teacher Model for Effective Knowledge Distillation} 

\titlerunning{MSE Teacher for Knowledge Distillation}

\author{Shayan Mohajer Hamidi$^*$\inst{1}\orcidlink{0000-0001-8321-7130} \and
Xizhen Deng$^*$\inst{2}\orcidlink{0009-0008-7974-848X} \and
Renhao Tan\inst{1}\orcidlink{0009-0005-6448-9954} \and Linfeng Ye\inst{1}\orcidlink{0009-0009-2355-1773} \and Ahmed Hussein Salamah\inst{1}\orcidlink{0000-0001-7836-5278
}}

\authorrunning{Sh. Mohajer Hamidi et al.}

\institute{University of Waterloo, Waterloo  ON N2L 3G1, Canada \email{\{smohajer,cameron.tan,l44ye,ahamsalamah\}@uwaterloo.ca}\and
University of Michigan,  Ann Arbor, MI 48109, USA
\email{xizhen@umich.edu}\\
$^*$ Authors contributed equally}

\maketitle

\begin{abstract}
Recently, it was shown that the role of the teacher in knowledge distillation (KD) is to provide the student with an estimate of the true Bayes conditional probability density (BCPD). Notably, the new findings propose that the student's error rate can be upper-bounded by the mean squared error (MSE) between the teacher's output and BCPD. Consequently, to enhance KD efficacy, the teacher should be trained such that its output is close to BCPD in MSE sense. This paper elucidates that training the teacher model with MSE loss equates to minimizing the MSE between its output and BCPD, aligning with its core responsibility of providing the student with a BCPD estimate closely resembling it in MSE terms. In this respect, through a comprehensive set of experiments, we demonstrate that substituting the conventional teacher trained with cross-entropy loss with one trained using MSE loss in state-of-the-art KD methods consistently boosts the student's accuracy, resulting in improvements of up to 2.6\%. The code for this paper is publicly available at: \url{https://github.com/ECCV2024MSE/ECCV_MSE_Teacher}.
  \keywords{Knowledge distillation \and Bayes conditional probability density \and Mean squared error}
\end{abstract}

\section{Introduction} \label{sec:intro}
Knowledge distillation (KD), as introduced by \cite{buciluǎ2006model} and popularized by \cite{hinton2015distilling}, has emerged as a highly effective model compression technique, and has received significant attention from both academia and industry in recent years. At its core, KD entails the process of transferring the knowledge of a cumbersome model (teacher) into a lightweight counterpart (student). After the pioneering work by \cite{hinton2015distilling}, numerous researchers have attempted to improve the performance of KD \cite{romero2014fitnets,anil2018large,park2019relational}, and to understand why distillation works \cite{ye2024bayes,phuong2019towards,mobahi2020self,ye2024bayes,allen2020towards,menon2021statistical,10487854,dao2020knowledge}.

An aspect of KD that has received relatively limited attention is the training of the teacher model. In most of the existing KD methods, the teacher is typically trained to optimize its own performance, despite the fact that such optimization does not necessarily translate into enhanced student performance \cite{cho2019efficacy,mirzadeh2020improved}. Hence, to effectively train a student to attain high performance, it is crucial to align the training of the teacher accordingly.

Recently, \cite{menon2021statistical} showed that the teacher's soft predictions can act as a proxy for the unknown true Bayes conditional probability distribution (BCPD) of label $y$ given an input $\boldsymbol{x}$. Specifically, a teacher model trained using conventional cross-entropy (CE) loss function approximates the true BCPD of the underlying dataset \cite{kanaya1991bayes}, and then it passes this estimate to the student model. As such, the enhancement in the student's accuracy within KD stems from the fact that the student utilizes the teacher's BCPD approximation to train its own model. Additionally, \cite{menon2021statistical} noted that as the teacher's predictions approach the BCPD, the generalization error of the student model decreases. More importantly, \cite{dao2020knowledge} showed that the classification error rate of the student is directly bounded by the MSE between the teacher's output and the BCPD, as established through the Rademacher analysis. This fact was further empirically confirmed by \cite{ren2021better} where they treated the teacher's output as a supervisory signal for the student model. Hence, based on these findings, it is imperative for effective KD that the teacher is trained such that its output is close to the BCPD in the MSE sense.

Now the question is that how the teacher model should be trained to ensure that its prediction is close to the BCPD in the MSE sense? To answer to this question, in this paper, we prove that training a DNN via MSE (resp. CE) loss is equivalent to training it to minimize the expected MSE (resp. CE) between its output and the true BCPD. Hence, based on the discussions above, for an effective KD, the teacher should be trained using MSE loss function. We shall note that proximity in terms of MSE does not necessarily equate to proximity in terms of CE, and vice versa. Therefore, while the output of a teacher trained using CE loss serves as an estimate of the true BCPD, it may not necessarily be close to the BCPD in terms of MSE, which is essential for effective KD. In fact, we empirically show that although the student's accuracy is almost inversely proportional to the MSE between BCPD and teacher's output, such relationship does not exist between the student's accuracy and CE between BCPD and teacher's output. 

Based on the discussions above, we claim that for an effective KD, the teacher should be trained using MSE loss. To demonstrate the effectiveness of the teacher trained via MSE loss in KD, we conduct a thorough set of experiments over CIFAR-100 and ImageNet datasets, and show that by \underline{\textbf{solely}} replacing a conventional teacher trained with CE loss by one trained with MSE loss in the existing state-of-the-art KD methods, the student's accuracy consistently increases. We shall emphasize the fact that such gain is obtained without making any changes over the underlying KD methods such as its distillation loss function or any hyper-parameters. Additionally, we observe a slight decrease in the teacher's performance when trained using MSE loss, confirming that optimizing the teacher's performance is not necessary for an effective knowledge distillation process. To summarize, the contributions of this paper are as follows
\begin{itemize}
    \item We introduce a theorem to show that training a DNN to minimize CE/MSE loss is equivalent to training it to minimize the CE/MSE of its output to the true BCPD.
    \item We show that for an effective KD, the student should be provided with an estimate of BCPD that is close to it in MSE sense; thus, as per our theorem, the teacher as a BCPD estimator should be trained via MSE loss.  
    \item We conduct a thorough set of experiments over CIFAR-100 and ImageNet datasets, and show that by replacing the teacher trained by CE with the one trained by MSE in the existing state-of-the-art KD methods, the student's accuracy consistently increases. 
\end{itemize}

\section{Related works} \label{sec:related}
\subsection{Knowledge distillation}
The concept of knowledge transfer, as a means of compression, was first introduced by \cite{buciluǎ2006model}. Then, \cite{hinton2015distilling} popularized this concept by softening the teacher's and student's logits using temperature technique where
the student mimics the soft probabilities of the teacher, and referred to it as KD. To improve the effectiveness of distillation, various forms of knowledge transfer methods have been introduced which could be mainly categorized into three types: (i) logit-based~\cite{zhu2018knowledge,stanton2021does,chen2020online,li2020online,beyer2022knowledge,DKD}, (ii) representation-based~\cite{romero2014fitnets, ATKD, yim2017gift, HSAKD}, and (iii) relationship-based~\cite{park2019relational,liu2019knowledge, CCKD,yang2022cross}.  

\subsection{Training a customized teacher for KD}
In the literature, only a few works trained teachers specifically tailored for KD. \cite{yang2019training} attempted to train a tolerant teacher which provides more secondary information to the student. They realized this via adding an extra term to 
facilitating a few secondary classes to emerge to complement the primary class. \cite{cho2019efficacy} regularized the teacher utilizing early-stopping during the training. Nevertheless, achieving optimal results may necessitate a thorough hyperparameter search, as the epoch number for identifying the best checkpoint can be particularly sensitive to various training settings, such as the learning rate schedule. Additionally, it is feasible to save multiple early teacher checkpoints, allowing the student to be distilled from them sequentially \cite{jin2019knowledge}.

In addition, \cite{wang2022efficient} stated that a checkpoint in the middle of the training procedure, often serves as a better teacher compared to the fully converged model. The authors in \cite{dong2023toward} used Lipschitz regularization so that the teacher can can learn the label distribution of the underlying dataset. Also \cite{ye2024bayes} trained the teacher to have high conditional mutual information so that it can better predict the true BCPD. \cite{tan2022improving} trained the teachers to have more dispersed soft probabilities.

\subsection{Training DNNs using MSE}
Mean squared error (MSE) loss serves as a prevalent choice for training DNNs, particularly in the context of regression tasks. This loss function is widely embraced when the objective is to predict continuous values, and its formulation involves calculating the average of the squared differences between predicted and actual values. Despite its established efficacy in regression scenarios, the landscape of loss functions in the realm of DNNs is vast and dynamic.

In contemporary practices, DNNs dedicated to classification tasks predominantly leverage the CE loss function. This method has gained substantial empirical favor, often surpassing MSE in the context of classification-oriented objectives. However, the empirical superiority of CE over MSE remains a topic of ongoing exploration. Notably, the existing body of literature does not uniformly advocate for a distinct advantage of CE in all scenarios.

Recent insights, as highlighted by the study conducted by \cite{hui2020evaluation}, challenge the prevailing notion by showcasing that models trained with MSE not only hold their ground against their CE-trained counterparts across a diverse spectrum of tasks and settings but, intriguingly, exhibit superior classification performance in the majority of experimental conditions. These findings prompt a reevaluation of the perceived hierarchy between MSE and CE, underscoring the need for nuanced considerations when selecting the most suitable loss function based on the specificities of the task at hand. In light of such empirical observations, the applicability and performance of MSE in DNN training extend beyond the traditional confines of regression, warranting a more comprehensive exploration of its utility across various domains and applications.

\section{Notation and Preliminaries}
\subsection{Notation}
For a positive integer $C$,  let $[C]\triangleq \{1,\dots,C\}$. We use bold lowercase letters (e.g., $\boldsymbol{p}$) to represent vectors. Denote by $\boldsymbol{p}[i]$ the $i$-th element of vector $\boldsymbol{p}$. Also, $\{ \boldsymbol{p}[c] \}_{c \in \mathcal{C}}$ is the set of all components of $\boldsymbol{p}$ with indices from the set $\mathcal{C}$. For two vectors $\boldsymbol{u}$ and $ \boldsymbol{v}$, denote by $\boldsymbol{u}\cdot \boldsymbol{v}$ their inner product. We use $|\mathcal{C}|$ to denote the cardinality of a set $\mathcal{C}$. The transpose operation is denoted by $(\cdot)^{\rm{T}}$.

The cross-entropy of two probability distributions $\boldsymbol{p}_1$ and $\boldsymbol{p}_2$ is defined as $H (\boldsymbol{p}_1, \boldsymbol{p}_2) = \sum_{c=1}^C -\boldsymbol{p}_1 [c] \log \boldsymbol{p}_2 [c]$. For a random variable $x$, denote by $\mathbb{P}_{x}$ its probability distribution, and by $\mathbb{E}_{x}[\cdot]$ the expected value operation w.r.t. $x$. For two random variables $x$ and $y$, denote by $\mathbb{P}_{(x,y)}$ their joint distribution.

\subsection{True risk Vs. empirical risk}
In a classification task with $C$ classes, a DNN could be regarded as a mapping $f_{\boldsymbol{\theta}}: \boldsymbol{x} \to \boldsymbol{p}_{\boldsymbol{x}}$, where $\boldsymbol{\theta}$ represents all the model parameters, $\boldsymbol{x} \in \mathbb{R}^d$ is an input image, and $\boldsymbol{p}_{\boldsymbol{x}} \in \Delta^C$, where $\Delta^C$ is the C dimensional probability simplex. Then, the classifier predicts the correct label of $\boldsymbol{x}$, denoted by $y$, as $\hat{y}=\arg \max_{c \in [C]} \boldsymbol{p}_{\boldsymbol{x}}[c]$. As such, the error rate of $f$ is defined as $\epsilon=\text{Pr} \{\hat{y} \neq y\}$, and its accuracy is equal to $1-\epsilon$. One may learn such a classifier by minimizing the \textit{true} risk 
\begin{align} \label{eq:risk}
R(f_{\boldsymbol{\theta}}, \ell) & \triangleq \mathbb{E}_{(\boldsymbol{x},y)} \left[ \ell \left( y, \boldsymbol{p}_{\boldsymbol{x}} \right) \right] = \mathbb{E}_{\boldsymbol{x}} \left[ \mathbb{E}_{y|\boldsymbol{x}} \left[ \ell \left( y, \boldsymbol{p}_{\boldsymbol{x}} \right)\right] \right] \nonumber \\ &=\mathbb{E}_{\boldsymbol{x}}  \left[ (\boldsymbol{p}^*_{\boldsymbol{x}})^{\rm{T}} \cdot \boldsymbol{\ell} ( \boldsymbol{p}_{\boldsymbol{x}} )\right],
\end{align}
where $\ell(\cdot)$ is the loss function and $\boldsymbol{\ell} ( \cdot ) \triangleq  [ \ell( 1, \cdot ), \ldots, \ell( C, \cdot ) ] $ is the vector of loss function, and $\boldsymbol{p}^*_{\boldsymbol{x}} \triangleq  \left[ \text{Pr}(y|\boldsymbol{x}) \right]_{y \in [C]}$ is Bayes class probability distribution over the labels, i.e, the BCPD.

However, in a typical deep learning algorithm, both the probability density function of $\boldsymbol{x}$, namely $\mathbb{P}_{\boldsymbol{x}}$, and also $\boldsymbol{p}^*_{\boldsymbol{x}}$ are unknown. Hence, one may learn such a classifier by instead minimizing the \emph{empirical} risk on a training sample $\mathcal{D} \triangleq \{ ( \boldsymbol{x}_n, y_n ) \}_{n = 1}^N$ defined as:
\begin{equation}
    \label{eqn:empirical-risk-one-hot}
    R_{\rm emp}(f_{\boldsymbol{\theta}}, \ell) \triangleq 
    \frac{1}{N} \sum_{n \in [N]} \mathbf{y}_n^{\rm{T}} \cdot \boldsymbol{\ell} \left( \boldsymbol{p}_{\boldsymbol{x}_n} \right),
\end{equation}
where $\mathbf{y}_n^{\rm{T}}$ is the one-hot vector with its $y_n$-th entry set to one and all other entries set to zero. By comparing \cref{eq:risk} and \cref{eqn:empirical-risk-one-hot}, we see that in \cref{eqn:empirical-risk-one-hot}: (i) 
$\mathbb{P}_{\boldsymbol{x}}$ is approximated by $\frac{1}{N}$, and (ii) $\boldsymbol{p}^*_{\boldsymbol{x}}$ is approximated by $\mathbf{y}$ which is an unbiased estimation of $\boldsymbol{p}^*_{\boldsymbol{x}}$. The former assumption is reasonable, however, the latter results in a significant loss in granularity. To delve deeper into this matter, it is crucial to recognize that images inherently carry a wealth of information, and the practice of assigning a one-hot vector to $\boldsymbol{p}^*_{\boldsymbol{x}}$ tends to lead to a significant loss of this information. As we will explore further in the subsequent subsection, KD emerges as a strategy that, to some extent, alleviates this issue, providing a mechanism to better transfer the nuanced information embedded within images.

\subsection{Estimating BCPD by the teacher in KD} \label{sec:KDloss}
In KD, the role of the teacher is to provide the student with a better estimate of $\boldsymbol{p}^*_{\boldsymbol{x}}$ compared to one-hot vectors. To elucidate, denote by $\boldsymbol{p}^t_{\boldsymbol{x}}$ and $\boldsymbol{p}^s_{\boldsymbol{x}}$ the pre-trained teacher's and student's outputs to sample $\boldsymbol{x}$, respectively. Then, the student uses the teacher's estimate of $\boldsymbol{p}^*_{\boldsymbol{x}}$, and minimizes
\begin{align} \label{eq:emp-student}
    R_{\rm kd}( f_{\boldsymbol{\theta}}, \ell) \triangleq 
    \frac{1}{N} \sum_{n \in [N]} \left( \boldsymbol{p}^t_{\boldsymbol{x}_n} \right)^{\rm{T}} \cdot \boldsymbol{\ell} \left(\boldsymbol{p}^s_{\boldsymbol{x}_n}\right).    
\end{align}
Note that the one-hot vector $\mathbf{y}_n$ in \cref{eqn:empirical-risk-one-hot} is now replaced by the teacher's output probability $\boldsymbol{p}^t_{\boldsymbol{x}_n}$ in \cref{eq:emp-student}. In fact, the effectiveness of KD lies in the fact that $\boldsymbol{p}^t_{\boldsymbol{x}_n}$ serves as a better estimate of $\boldsymbol{p}^*_{\boldsymbol{x}}$ compared to the one-hot vector $\mathbf{y}_n^{\rm{T}}$.

\subsection{Student's generalization error and accuracy}\label{sec:generr}
In this subsection, our objective is to identify the key characteristics that the estimated BCPD should possess in order to enhance the accuracy of the student.

As shown by \cite{menon2021statistical}, the student's generalization error is upper-bounded as
\begin{align} \label{eq:empVstrue}
\mathbb{E} \left[ (R_{\rm kd}( f_{\boldsymbol{\theta}}, \ell) - R(f_{\boldsymbol{\theta}}, \ell))^2 \right]  \leq \frac{1}{N} \text{Var} \left[ (\boldsymbol{p}^t_{\boldsymbol{x}})^{\rm{T}} \cdot \boldsymbol{\ell} (\boldsymbol{p}_{\boldsymbol{x}}) \right]   
+ \kappa \left( \mathbb{E} \left[ \|\boldsymbol{p}^t_{\boldsymbol{x}} - \boldsymbol{p}^*_{\boldsymbol{x}} \| \right] \right)^2,
\end{align}
where $\kappa$ is a positive constant number. 
When $N$ is large, which is commonly the case for datasets in existing literature, the second term will dominate the right-hand side of \cref{eq:empVstrue}. This implies that smaller average $\|\boldsymbol{p}^t_{\boldsymbol{x}} - \boldsymbol{p}^*_{\boldsymbol{x}}\|_2$ will lead to $R_{\rm emp}(f_{\boldsymbol{\theta}})$ being a better approximation of the true risk $R(f_{\boldsymbol{\theta}})$, minimizing it should then lead to a better learned model. 

On the other hand, \cite{ren2021better} empirically showed that the accuracy of the student is almost inversely proportional to $\|\boldsymbol{p}^t_{\boldsymbol{x}} - \boldsymbol{p}^*_{\boldsymbol{x}}\|$. In addition, \cite{dao2020knowledge} showed that the accuracy of the student is directly bounded by the MSE
between teacher’s prediction and BCPD through the Rademacher analysis.

Therefore, the quality of the estimates provided by the teacher to the student can be measured by the MSE between its output and to the true $\boldsymbol{p}^*_{\boldsymbol{x}}$. Based on this, in the next section, we show that for an effective KD, the teacher should be indeed trained via MSE loss, and not CE loss. 

\section{Methodology} \label{sec:meth}
In this section, first in \cref{sec:mse-ce}, we introduce a theorem demonstrating that training a DNN model with MSE (CE) loss function results in minimizing the MSE (CE) between its output and the BCPD. Then, in \cref{sec:mse-ce2}, we use a synthetic dataset to empirically validate the introduced theorem, and to show that (i) closeness in MSE sense does not necessarily mean closeness in CE sense, and (ii) for an effective KD the teacher's output should be close to the true BCPD in MSE sense. Then, we conclude that the teacher should be trained via MSE loss function.

\subsection{MSE loss Vs. CE loss} \label{sec:mse-ce}
In a classification task with $C$ classes, it has been shown that the risk in \cref{eq:risk}, for $\ell= \{ \text{CE}, \text{MSE}\}$, is minimized when $\boldsymbol{p}_{\boldsymbol{x}}=\boldsymbol{p}^*_{\boldsymbol{x}}$ \cite{kanaya1991bayes}. However, since the underlying $\mathbb{P}_{(\boldsymbol{x},y)}$ is unknown, the DNNs are trained to minimize the empirical risk in \cref{eqn:empirical-risk-one-hot}, and consequently they can only approximate the true BCPD\footnote{Aside from the fact that $\mathbb{P}_{(\boldsymbol{x},y)}$ is unknown, training cannot typically find the global minimum hindering the DNNs to give accurate BCPD.}.

Hence, a teacher trained by either CE or MSE can approximate the true BCPD. However, it is crucial to note that these two estimates differ, as precisely established in the following theorem.

\begin{theorem}\label{th:eq}
For $\ell= \{ \text{CE}, \text{MSE}\}$
\begin{align}
\min_{\boldsymbol{\theta}}  \mathbb{E}_{(\boldsymbol{x},y)} \left[ \ell \left( y, \boldsymbol{p}_{\boldsymbol{x}} \right) \right] 	\equiv \min_{\boldsymbol{\theta}} \mathbb{E}_{(\boldsymbol{x},y)} \left[ \ell \left( \boldsymbol{p}^*_{\boldsymbol{x}}, \boldsymbol{p}_{\boldsymbol{x}} \right) \right].
\end{align}
\end{theorem}

\begin{proof}
Please refer to the \textit{Supplementary materials.}     
\end{proof}

\cref{th:eq} implies that when minimizing the expected loss, the resulting model attempts to generate outputs that closely approximate those of an "ideal" model. More importantly, this degree of closeness is quantified by the loss function. Specifically, (i) if $\ell=\text{MSE}$ then a model trained to minimize the expected squared error between $y$ and $\boldsymbol{p}_{\boldsymbol{x}}$ will generate outputs that minimize the expected squared error between $\boldsymbol{p}^*_{\boldsymbol{x}}$ and $\boldsymbol{p}_{\boldsymbol{x}}$; and (ii) if $\ell=\text{CE}$ then a model trained to minimize the expected cross-entropy between $y$ and $\boldsymbol{p}_{\boldsymbol{x}}$ will generate outputs that minimize the expected cross-entropy between $\boldsymbol{p}^*_{\boldsymbol{x}}$ and $\boldsymbol{p}_{\boldsymbol{x}}$. 

Thus, the BCPD estimates provided by the teachers trained by CE loss and MSE loss are different; in that, the former estimate is close to the true BCPD in CE sense, also the latter estimate is close to the true BCPD in MSE sense. In the next subsection, we empirically show that for the student to have high accuracy, the teacher's estimate should be close to the true BCPD in MSE sense, rather than in CE sense.

\subsection{MSE proximity Vs. CE proximity} \label{sec:mse-ce2}

In this subsection, our intention is two-fold: (i) we aim to empirically show that although the student's accuracy is almost inversely proportional to the MSE between BCPD and teacher's output (as also demonstrated by \cite{ren2021better,dao2020knowledge}), such relationship does not exist between the student's accuracy and CE between BCPD and teacher's output; and (ii) to empirically validate \cref{th:eq}. Toward this aim, since the true BCPD is unknown for the popular datasets in the literature, we generate a synthetic dataset whose BCPD is known. 

\noindent $\bullet$ \textbf{Generating dataset}: Inspired from \cite{ren2021better}, we generate a 3-class toy Gaussian dataset with $10^5$ data points. The dataset is divided into training, validation, and test sets with a split ratio $[0.9, 0.05, 0.05]$. 

The sampling process is implemented as follows: we first choose the label $y$ using a uniform distribution across all the 3 classes. Next, we sample $\left.x\right|_{y=k} \sim \mathcal{N}\left(\mu_k, \sigma^2 I\right)$ as the input signal. Here, $\mu_k$ is a 30-dim vector with entries randomly selected from $\{-\delta_\mu, 0, \delta_\mu\}$. Then, we calculate the BCPD of the samples using the fact that $p^*(y|\boldsymbol{x})\propto p(\boldsymbol{x}|y)p(y)$. Particularly, as $y$ follows uniform distribution, we have $\boldsymbol{p}^*(y|\boldsymbol{x})=\frac{\boldsymbol{p}(\boldsymbol{x}|y=k)}{\sum_{j\neq k}p(\boldsymbol{x}|y=j)}$. Following $\boldsymbol{p}(\boldsymbol{x}|y=k)\sim \mathcal{N}(\mu_k,\sigma^2I)$, we find $\boldsymbol{p}^*(y|\boldsymbol{x})$ should have a Softmax form as
\begin{equation}
    \boldsymbol{p}^*(y=k|\boldsymbol{x})=\frac{\text{exp} (s_k)}{\sum_{j\neq k}\text{exp} (s_j)};\quad s_i = -\frac{1}{2\sigma^2}\|\boldsymbol{x}-\mu_i\|^2_2.
\end{equation}

\noindent $\bullet$ \textbf{Training setup}: The underlying model (for both teacher and student) is a 3-layer MLP with ReLU activation function, and the hidden size is 128 for each layer. We set the learning rate as $5\times10^{-4}$, the batch size as $32$, and the number of training epochs is $100$. In our experiments, we set $\sigma=4$ and  $\delta_\mu=1$.

Now, we conduct two sets of experiments as elaborated in the sequel.

\subsubsection{Set 1: the student's accuracy as a function of MSE($\boldsymbol{p}^*_{\boldsymbol{x}}$, $\boldsymbol{p}_{\boldsymbol{x}}$) and CE($\boldsymbol{p}^*_{\boldsymbol{x}}$, $\boldsymbol{p}_{\boldsymbol{x}}$).}
\begin{figure*}[!t]
\centering  \includegraphics[width=1\textwidth]{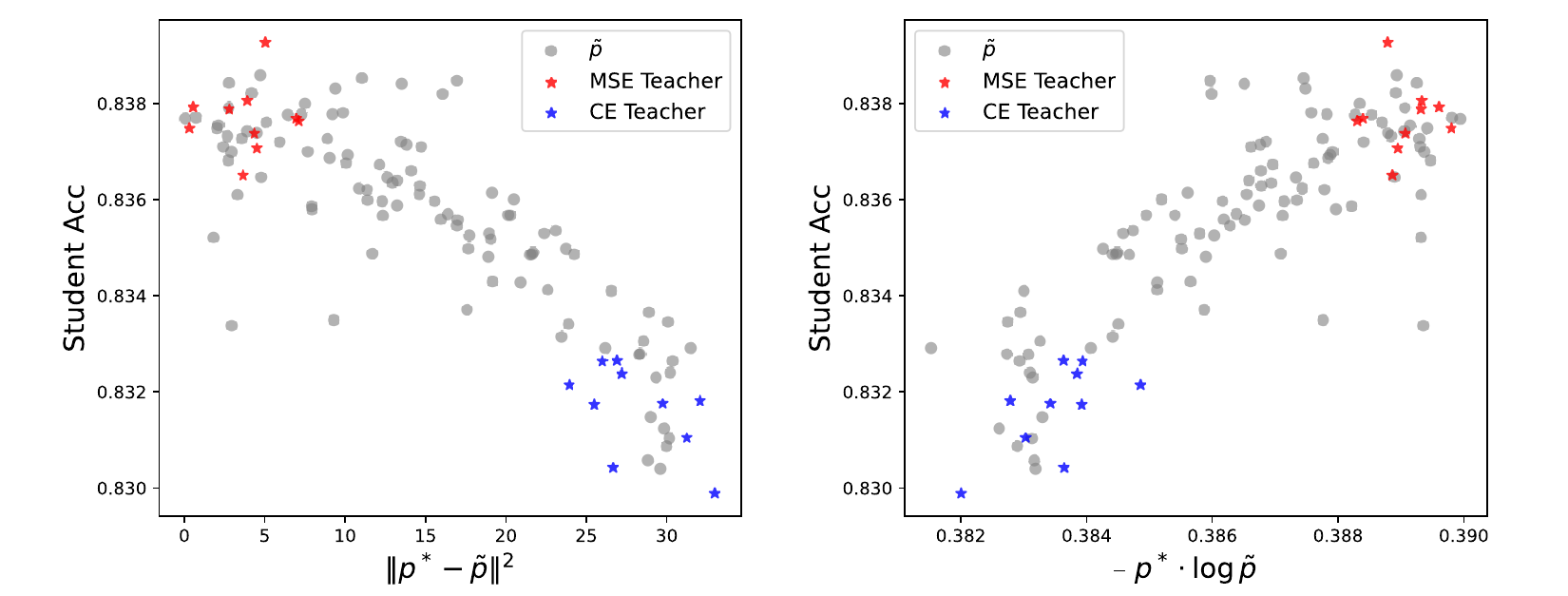}
  \caption{Student accuracy as a function of (left) the MSE between $\boldsymbol{p}_{\boldsymbol{x}}$ and $\Tilde{\boldsymbol{p}}_{\boldsymbol{x}}$; and (right) CE between $\boldsymbol{p}_{\boldsymbol{x}}$ and $\Tilde{\boldsymbol{p}}_{\boldsymbol{x}}$. The gray dots are noisy versions of the true $\boldsymbol{p}_{\boldsymbol{x}}$. Also, the red and blue dots represent the points corresponding to the estimates provided by MSE and CE teachers, respectively. 
} 
\label{fig:con_sep}
  \vspace{-0.1in}
\end{figure*}

Here, we conduct some experiments to understand which type of proximity, i.e., whether proximity in CE sense or MSE sense, is better for the sake of the student accuracy. To this end, we generate 100 perturbed/noisy versions of $\boldsymbol{p}^*_{\boldsymbol{x}}$ by adding some random noise to it, and denote any such perturbed version by $\Tilde{\boldsymbol{p}}^*_{\boldsymbol{x}}$. Then, we train the student 100 times, each time using one of the noisy $\Tilde{\boldsymbol{p}}^*_{\boldsymbol{x}}$, i.e., each time the student is trained via loss \cref{eq:emp-student} with $\boldsymbol{p}^t_{\boldsymbol{x}_n}$ replaced by $\Tilde{\boldsymbol{p}}^*_{\boldsymbol{x}}$. The resulting student's accuracy corresponding to these 100 noisy BCPD is depicted in \cref{fig:con_sep}, where in the left and right figure we set x-axis to the MSE and CE between the $\boldsymbol{p}^*_{\boldsymbol{x}}$ and $\Tilde{\boldsymbol{p}}^*_{\boldsymbol{x}}$, respectively (the points corresponding to noisy $\boldsymbol{p}^*_{\boldsymbol{x}}$ are denoted by \underline{\textbf{\textcolor{gray}{gray}}} dots). 

As seen in \cref{fig:con_sep}, the student's accuracy is almost inversely proportional to the MSE between $\boldsymbol{p}^*_{\boldsymbol{x}}$ and $\Tilde{\boldsymbol{p}}^*_{\boldsymbol{x}}$. Consequently, the teacher can enhance the student's performance by providing it with an estimate of $\boldsymbol{p}^*_{\boldsymbol{x}}$ that is closer to it in MSE sense. On the other hand, as seen in the right figure depicted in \cref{fig:con_sep}, such relationship does not exist; in that, minimizing the CE between $\boldsymbol{p}^*_{\boldsymbol{x}}$ and $\Tilde{\boldsymbol{p}}^*_{\boldsymbol{x}}$ does not necessarily increase the student's performance.

\subsubsection{Set 2: empirically validating Theorem 1.} The previous set of experiments showed that the student's accuracy can be improved if it is given a BCPD estimate which is close to $\boldsymbol{p}^*_{\boldsymbol{x}}$ in MSE sense. Here, we aim to empirically validate the \cref{th:eq}, and therefore based on our discussions above, the teacher should be trained via MSE loss. 

Hence, we train the teacher 10 times using MSE loss, and 10 times using CE loss. We denote the teacher's estimate of $\boldsymbol{p}^*_{\boldsymbol{x}}$ by $\Tilde{\boldsymbol{p}}^*_{\boldsymbol{x},\text{MSE}}$ and  $\Tilde{\boldsymbol{p}}^*_{\boldsymbol{x},\text{CE}}$ when it is trained by CE and MSE losses, respectively. 

Now, once again we train the student using these two types of estimate (10 estimates for each type) and record the student's accuracy. The results are depicted in \cref{fig:con_sep}, where we used red and blue dots to show the points corresponding to $\Tilde{\boldsymbol{p}}^*_{\boldsymbol{x},\text{MSE}}$ and  $\Tilde{\boldsymbol{p}}^*_{\boldsymbol{x},\text{CE}}$ estimates, respectively. 

Observing the two figures we can conclude that (i) training the teacher using MSE/CE loss yields the BCPD estimate which is close to $\boldsymbol{p}^*_{\boldsymbol{x}}$ in MSE/CE sense; (ii) the closeness in CE and MSE sense is rather different; and (iii) the teacher trained via MSE loss results in a better performance for the student model.

Lastly, we provide a toy example to understand  why proximity in MSE sense is different from that in CE sense. Assume that the true BCPD is $\boldsymbol{p}^*= [0.3, 0.7]$. Consider the following two estimates of the BCPD. 

\begin{itemize}
    \item Estimate 1, $E_1= [0.29, 0.71]$. Here $\text{MSE}(\boldsymbol{p}^*, E_1) = 0.01$ and $\text{CE}(\boldsymbol{p}^*, E_1) =  0.610$.

    \item Estimate 2, $E_2= [0.2, 0.8]$. Here $\text{MSE}(\boldsymbol{p}^*, E_2) = 0.10$ and $\text{CE}(\boldsymbol{p}^*, E_2) =  0.51$.
\end{itemize}

In the CE sense, $E_2$ is a superior estimator of $\boldsymbol{p}^*$. However, from the standpoint of MSE, $E_1$ is a more accurate representation of $\boldsymbol{p}^*$. Consequently, if a DNN is trained using the CE loss, it tends to produce an output akin to $E_2$ which is at a relatively greater distance from the true distribution $\boldsymbol{p}^*$ in the MSE sense.

\vspace{5mm}
\section{Experiments} \label{sec:Experiments}
We conclude the paper with extensive experiments demonstrating the superior effectiveness of MSE teacher compared to CE teacher in different state-of-the-art KD methods. The outcomes of these experiments collectively contribute to a compelling argument for the preferential utilization of MSE teachers in the KD variants.

\noindent $\bullet$  \textbf{Terminology}: Hereafter, we refer to teachers trained by the CE and MSE losses as the ``CE teacher" and ``MSE teacher", respectively.

\noindent $\bullet$  \textbf{Organization}:
The experiments are organized as follows: in \cref{sec:cifar} and \cref{sec:cifar2}, we conduct experiments on CIFAR-100 dataset; also \cref{sec:imag} provides experiments on ImageNet dataset. In \cref{sec:semi}, we compare MSE and CE teachers in the semi-supervised distillation task. Lastly, we evaluate the performance of MSE teacher in binary classification knowledge distillation task in \cref{sec:binary}.

\noindent $\bullet$ \textbf{Plug-and-play nature of MSE teacher}: In all the experiments conducted in this section, when evaluating the performance of the MSE teacher, we refrain from tuning any hyper-parameters in the underlying knowledge transfer methods, i.e., all hyper-parameters remain the same as those used in the corresponding benchmark methods. We present the classification accuracies on the test set for both the teacher model and its distilled student model.

\begin{table*}[h!]
\begin{center}
\caption{The test accuracy $(\%)$ of student networks on CIFAR-100 (averaged over 5 runs), with teacher-student pairs of the same/different architectures. The subscript denotes the improvement achieved by replacing CE teacher with MSE teacher. We use \textbf{bold} numbers and asterisk $(^*)$ to denote the best results and to identify the results reproduced on our local machines, respectively.} 
\vskip -0.1in
\resizebox{1\textwidth}{!}{
\begin{tabular}{ccc|cc|cc|cc|cc|cc} 
\toprule
\rowcolor{mygray} \multicolumn{13}{c}{~~~~~~~~Teachers and students with the \textbf{same} architectures.}
\\
\bottomrule
\bottomrule
 \multirow{2}{*}{Teacher} 
        & \multicolumn{2}{c|}{ResNet-56} & \multicolumn{2}{c|}{ResNet-110} & \multicolumn{2}{c|}{ResNet-110}   & \multicolumn{2}{c|}{WRN-40-2}  & \multicolumn{2}{c|}{WRN-40-2} & \multicolumn{2}{c}{VGG-13}     \\
        & CE      & MSE                & CE         & MSE                & CE       & MSE                  & CE     & MSE                 & CE     & MSE                & CE       & MSE              \\ 
 Accuracy& 72.34    & 72.14                 & 74.31     & 73.43               & 74.31     & 73.43                 & 75.61   &74.99                & 75.61   & 74.99               & 74.64     & 73.20   \\ \hline
Student & \multicolumn{2}{c|}{ResNet-20}   & \multicolumn{2}{c|}{ResNet-20}  & \multicolumn{2}{c|}{ResNet-32}    & \multicolumn{2}{c|}{WRN-16-2}  & \multicolumn{2}{c|}{WRN-40-1} & \multicolumn{2}{c}{VGG-8}      \\ 
Accuracy& \multicolumn{2}{c|}{69.06}       & \multicolumn{2}{c|}{69.06}      & \multicolumn{2}{c|}{71.14}        & \multicolumn{2}{c|}{73.26}     & \multicolumn{2}{c|}{71.98}    & \multicolumn{2}{c}{70.36}      \\ \hline
AT      & 70.55    & 70.80                 & 70.22     & 70.58               & 72.31     & 73.50                 & 74.08   & 74.30                & 72.77   & 73.05               & 71.43     & 71.72              \\[-0.4em]
        &          &\tiny ~~~${+0.25}$     &           &\tiny ~~~${+0.36}$   &           &\tiny~~~$+1.19$        &         &\tiny ~~~${+0.22}$    &         &\tiny~~~${+0.28}$    &           &\tiny ~~~${+0.29}$  \\
PKT     & 70.34    & 70.84                 & 70.25     & 70.55               & 72.61     & 72.90                 & 74.54   & 74.85                & 73.45   & 74.10               & 72.88     & 73.10              \\[-0.4em]
        &          &\tiny ~~~${+0.50}$     &           &\tiny ~~~${+0.30}$   &           & \tiny ~~~${+0.29}$    &         &\tiny ~~~${+0.31}$    &         &\tiny~~~${+0.65}$    &           &\tiny ~~~${+0.47}$  \\
SP      & 69.67    & 70.77                 & 70.04     & 70.75               & 72.69     & 73.34                 & 73.83   & 74.60                & 72.43   & 73.30               & 72.68     & 73.19              \\[-0.4em]
        &          &\tiny~~~$+1.10$        &           &\tiny ~~~${+0.71}$   &           & \tiny ~~~${+0.65}$    &         &\tiny ~~~$+0.77$      &         &\tiny~~~$+0.87$      &           &\tiny ~~~${+0.61}$  \\
CC      & 69.63    & 69.99                 & 69.48     & 69.89               & 71.48     & 71.75                 & 73.56   & 73.87                & 72.21   & 72.50               & 70.71     & 71.00              \\[-0.4em]
        &          &\tiny ~~~${+0.36}$     &           &\tiny ~~~${+0.41}$   &           & \tiny ~~~${+0.27}$    &         &\tiny ~~~${+0.31}$    &         &\tiny~~~${+0.29}$    &           &\tiny ~~~${+0.29}$  \\
RKD     & 69.61    & 70.50                 & 69.25     & 70.20               & 71.82     & 72.62                 & 73.35   & 73.66                & 72.22   & 72.67               & 71.48     & 71.88              \\[-0.4em]
        &          &\tiny ~~~${+0.89}$     &           &\tiny~~~$+0.95$      &           & \tiny ~~~${+0.80}$    &         &\tiny ~~~${+0.31}$    &         &\tiny~~~${+0.45}$    &           &\tiny ~~~${+0.40}$  \\
VID     & 70.38    & 70.61                 & 70.16     & 70.49               & 72.61     & 73.05                 & 74.11   & 74.44                & 73.30   & 73.58               & 71.23     & 71.57              \\[-0.4em]
        &          &\tiny ~~~${+0.23}$     &           &\tiny ~~~${+0.33}$   &           & \tiny ~~~${+0.44}$    &         &\tiny ~~~${+0.33}$    &         &\tiny~~~${+0.28}$    &           &\tiny ~~~${+0.34}$  \\
CRD     & 71.16    & 71.43                 & 71.46     & 71.87               & 73.48     & 74.03                 & 75.48   & 75.85                & 74.14   & 74.86               & 73.94     & 74.25              \\[-0.4em]
        &          &\tiny ~~~${+0.27}$     &           &\tiny ~~~${+0.41}$   &           & \tiny ~~~${+0.55}$    &         &\tiny ~~~${+0.37}$    &         &\tiny~~~${+0.72}$    &           &\tiny ~~~${+0.31}$  \\
REVIEW  & 71.89    & 72.15                 &~~71.65$^*$& 72.04             & 73.89     & 74.02                 & 76.12   & 76.29               & 75.09   & 75.33               & 74.84     & 74.90              \\[-0.4em]
        &          &\tiny ~~~${+0.26}$     &           &\tiny ~~~${+0.39}$   &           & \tiny ~~~${+0.13}$    &         &\tiny ~~~${+0.17}$    &         &\tiny~~~${+0.24}$    &           &\tiny ~~~${+0.06}$  \\ 
DKD     & 71.97    & 72.27                 &~~71.51$^*$& 71.88               & 74.11     & 74.32                 & 76.24   & 76.76                & 74.81   & 75.53               & 74.68     & 74.92              \\[-0.4em]
        &          &\tiny ~~~${+0.30}$     &           &\tiny ~~~${+0.37}$   &           & \tiny ~~~${+0.21}$    &         &\tiny ~~~${+0.52}$    &         &\tiny~~~${+0.72}$    &           &\tiny ~~~${+0.24}$  \\
HSAKD   & 72.58    & \textbf{72.74}        &~~72.64$^*$& \textbf{73.07}      &~~74.97$^*$& \textbf{75.52}        & 77.20   & \textbf{77.55}       & 77.00   & \textbf{77.32}      &~~75.42$^*$&\textbf{75.76}      \\[-0.4em]
        &          &\tiny ~~~${+0.16}$     &           & \tiny ~~~${+0.43}$  &           & \tiny ~~~${+0.55}$    &         &\tiny ~~~${+0.35}$    &         &\tiny ~~~${+0.32}$   &           & \tiny ~~~${+0.34}$ \\ 
\toprule
\rowcolor{mygray} \multicolumn{13} {c}{~~~~~~~~Teachers and students with \textbf{different} architectures.}
\\
\bottomrule
\bottomrule
 \multirow{2}{*}{Teacher} 
        & \multicolumn{2}{c|}{ResNet-50}   & \multicolumn{2}{c|}{ResNet-50}  & \multicolumn{2}{c|}{ResNet-32$\times$4}  & \multicolumn{2}{c|}{ResNet-32$\times$4}& \multicolumn{2}{c|}{WRN-40-2}    & \multicolumn{2}{c}{VGG-13}        \\
        & CE       & MSE                 & CE       & MSE                & CE       & MSE                         & CE     & MSE                         & CE     & MSE                   & CE       & MSE                   \\ 
Accuracy& 79.34     & 74.54                & 79.34     & 74.54               & 79.41     & 75.24                         & 79.41   & 75.24                         & 75.61   & 74.99                  & 74.64     & 73.20               \\\hline
Student & \multicolumn{2}{c|}{MobileNetV2} & \multicolumn{2}{c|}{VGG-8}      & \multicolumn{2}{c|}{ShuffleNetV1}        & \multicolumn{2}{c|}{ShuffleNetV2}      & \multicolumn{2}{c|}{ShuffleNetV1}& \multicolumn{2}{c}{MobileNetV2}  \\ 
Accuracy& \multicolumn{2}{c|}{64.60}       & \multicolumn{2}{c|}{70.36}      & \multicolumn{2}{c|}{70.50}               & \multicolumn{2}{c|}{71.82}             & \multicolumn{2}{c|}{70.50}       & \multicolumn{2}{c}{64.60}        \\ \hline
AT      & 58.58     & 59.63                & 71.84     & 72.12               & 71.73     & 72.06                        & 72.73   & 74.11                        & 73.32   & 74.33                  & 59.40     & 62.07                \\[-0.4em]
        &           &\tiny ~~~$+1.05$      &           & \tiny ~~~$+0.28$    && \tiny ~~~$+0.33$                        &         &\tiny ~~~$+1.38$              &         & \tiny ~~~$+1.01$       &           &\tiny ~~~$+2.67$\\
PKT     & 66.52     & 67.02                & 73.10     & 73.45               & 74.10     & 74.81                        & 74.69   & 76.34                        & 73.89   & 75.39                  & 67.13     & 68.08                \\[-0.4em]
        &           &\tiny ~~~$+0.50$      &           & \tiny ~~~$+0.35$    &           & \tiny ~~~$+0.71$             &         & \tiny ~~~$+1.65$             &         & \tiny ~~~$+1.50$       &           & \tiny ~~~$+0.95$     \\
SP      & 68.08     & 69.00                & 73.34     & 74.04               & 73.48     & 74.57                        & 74.56   & 75.70                        & 74.52   & 75.72                  & 66.30     & 67.03                \\[-0.4em]
        &           &\tiny ~~~$+0.92$      &           & \tiny ~~~$+0.70$    &           & \tiny ~~~$+1.09$             &         & \tiny ~~~$+1.14$             &         &\tiny~~~$+1.20$         &           & \tiny ~~~$+0.73$     \\
CC      & 65.43     & 65.90                & 70.25     & 70.90               & 71.14     & 71.77                        & 71.29   & 73.02                        & 71.38   & 71.80                  & 64.86     & 65.05                \\[-0.4em]
        &           &\tiny ~~~$+0.47$      &           & \tiny ~~~$+0.65$    &           & \tiny ~~~$+0.63$             &         & \tiny ~~~$+1.73$             &         & \tiny ~~~$+0.42$       &           & \tiny ~~~$+0.19$     \\
RKD     & 64.43     & 64.88                & 71.50     & 72.05               & 72.28     & 73.19                        & 73.21   & 73.62                        & 72.21   & 73.25                  & 64.52     & 65.32                \\[-0.4em]
        &           &\tiny ~~~$+0.45$      &           & \tiny ~~~$+0.55$    &           & \tiny ~~~$+0.91$             &         & \tiny ~~~$+0.41$             &         & \tiny ~~~$+1.04$       &           & \tiny ~~~$+0.80$     \\
VID     & 67.57     & 67.77                & 70.30     & 70.55               & 73.38     & 73.89                        & 73.40   & 74.67                        & 73.61   & 75.03                  & 65.56     & 65.82                \\[-0.4em]
        &           &\tiny ~~~$+0.20$      &           & \tiny ~~~$+0.25$    &           & \tiny ~~~$+0.51$             &         & \tiny ~~~$+1.27$             &         & \tiny ~~~$+1.42$       &           & \tiny ~~~$+0.26$     \\
CRD     & 69.11     & 69.15                & 74.30     & 74.55              & 75.11     & 75.81                        & 75.65   & 76.54                        & 76.05   & 76.43                  & 69.70     & 69.77                \\[-0.4em]
        &           &\tiny ~~~$+0.04$      &           & \tiny ~~~$+0.25$    &           & \tiny ~~~$+0.70$             &         & \tiny ~~~$+0.89$             &         & \tiny ~~~$+0.38$       &           & \tiny ~~~$+0.07$     \\ 
REVIEW  & 69.89     & 70.03               &~~73.43$^*$& 73.90               & 77.45     & 77.78                        & 77.78   & 78.81                        & 77.14   & 77.25                  & 70.37     & 70.81               \\[-0.4em]
        &           &\tiny ~~~$+0.14$      &           & \tiny ~~~$+0.47$    &           & \tiny ~~~$+0.33$             &         & \tiny ~~~$+0.03$             &         & \tiny ~~~$+0.11$       &           & \tiny ~~~$+0.44$     \\
DKD     & 70.35     & 71.40                &~~73.94$^*$& 75.14               & 76.45     & 77.15                        & 77.07   & 77.52                        & 76.70   & 77.30                  & 69.71     & 70.22                \\[-0.4em]
        &           &\tiny ~~~$+1.05$      &           &\tiny ~~~$+1.17$     &           & \tiny ~~~$+0.70$             &         & \tiny ~~~$+0.45$             &         & \tiny ~~~$+0.60$       &           & \tiny ~~~$+0.51$     \\
HSAKD   &~~71.83$^*$&\textbf{72.67}        &~~75.87$^*$& \textbf{76.34}      &~~79.51$^*$& \textbf{79.81}               & 79.93   & \textbf{80.09}               & 78.51   & \textbf{78.82}         &~~71.09$^*$&\textbf{72.40}        \\[-0.4em]
        &           &\tiny ~~~$+0.84$      &           & \tiny ~~~$+0.47$    &           & \tiny ~~~${+0.30}$           &         & \tiny ~~~${+0.17}$           &         & \tiny ~~~$+0.31$       &           & \tiny ~~~$+1.31$     \\
        \bottomrule
\end{tabular}}
\label{table:cifar100}
\end{center}
\end{table*}

In addition, we should note that for training the MSE teacher, we use the same training setup as that used for the CE loss. One may further improve our results by further tuning the training hyper-parameters. In addition, similarly to the conventional KD methods, we use the same teacher for different students, without adjusting the teacher based on the specifics of each student model.  

\vspace{5mm}

\subsection{CIFAR-100} \label{sec:cifar}
This dataset comprises 50,000 training and 10,000 test color images, each of size $32\times32$, and is annotated for 100 classes \cite{krizhevsky2009learning}. 

\noindent $\bullet$ \textbf{Teacher-student pairs:} 
In line with the configurations of CRD \cite{CRD}, we use a couple of teacher-student pairs with identical and different network architectures for our experiments (see \cref{table:cifar100}). We conduct each experiment over 5 independent runs and report the average accuracy (for the accuracy variances, refer to the \textit{Supplementary materials}).

\noindent $\bullet$ \textbf{KD variants:} 
For comprehensive comparisons, we compare using a MSE teacher Vs. CE teacher in the existing state-of-the-art distillation methods, including KD \cite{hinton2015distilling}, AT \cite{ATKD}, PKT \cite{PKTKD}, SP \cite{SPKD}, CC \cite{CCKD}, RKD \cite{RKD}, VID \cite{VID}, CRD \cite{CRD}, DKD \cite{DKD}, REVIEWKD \cite{REVIEWKD}, and HSAKD \cite{HSAKD}.

\noindent $\bullet$ \textbf{Training setup:} For all variants of knowledge distillation and settings in this paper, SGD is applied as the optimizer. We train the student for 240 epochs for all experiments with an initial learning rate of 0.05 by default, which will be decayed by factor of 0.1 at epoch 150, 180, 210. For MobileNetV2, ShuffleNetV1, and ShuffleNetV2, a smaller initial learning rate of 0.01 is used. We adopt batch size of 64. In addition, we report the hyper-parameters used for underlying KD variants in \textit{Supplementary materials}.

\begin{table}[ht] 
\begin{center}
\caption{Additional experiments on CIFAR-100, when the feature based distillation methods are combined with conventional KD method. The test accuracy $(\%)$ of student networks on CIFAR-100 (averaged over 5 runs), with teacher-student pairs of the same/different architectures. The subscript denotes the improvement achieved by replacing CE teacher with MSE teacher.} \label{table:cifar100_2}
\vskip -0.1in
\resizebox{1\textwidth}{!}{
\begin{tabular}{ccc|cc|cc|cc|cc|cc} 
\toprule
 \multirow{2}{*}{Teacher} 
        & \multicolumn{2}{c|}{ResNet-110} & \multicolumn{2}{c|}{WRN-40-2}  & \multicolumn{2}{c|}{VGG-13}  & \multicolumn{2}{c|}{ResNet-50}  & \multicolumn{2}{c|}{ResNet-32$\times$4}  & \multicolumn{2}{c}{VGG-13}        \\
        & CE      & MSE                & CE         & MSE                & CE       & MSE        & CE       & MSE                 & CE       & MSE                & CE       & MSE          \\ 
 Accuracy   & 74.31     & 73.43                             & 75.61   & 74.99                           & 74.64     & 73.20      & 79.34     & 74.54                           & 79.41   & 75.24                                  & 74.64     & 73.20        \\
 \hline
Student    & \multicolumn{2}{c|}{ResNet-20}     & \multicolumn{2}{c|}{WRN-16-2}  &  \multicolumn{2}{c|}{VGG-8}   & \multicolumn{2}{c|}{VGG-8}         & \multicolumn{2}{c|}{ShuffleNetV2}     & \multicolumn{2}{c}{MobileNetV2}   \\ 
Accuracy       & \multicolumn{2}{c|}{69.06}             & \multicolumn{2}{c|}{73.26}      & \multicolumn{2}{c|}{70.36}   & \multicolumn{2}{c|}{71.14}             & \multicolumn{2}{c|}{73.26}              & \multicolumn{2}{c}{64.60}   \\ \hline
AT+KD            & 70.97     & 71.34                       & 75.32   & 75.65                           & 73.48     & 73.89     & 74.01     & 74.22                            & 75.39   & 77.63                           & 65.13     & 66.76         \\[-0.4em]
             &           &\tiny ~~~${+0.37}$          &         &\tiny ~~~${+0.33}$    &           &\tiny ~~~${+0.41}$ &           & \tiny ~~~$+0.21$            &         &\tiny ~~~$+2.24$              &           &\tiny ~~~$+1.63$ \\
PKT+KD           & 70.72     & 71.15                        & 75.33   & 75.73                           & 73.25     & 73.50     & 73.61     & 73.80            & 74.66   & 76.13               & 68.13     & 68.89         \\[-0.4em]
            &           &\tiny ~~~${+0.43}$    &         &\tiny ~~~${+0.40}$     &           &\tiny ~~~${+0.25}$ &           & \tiny ~~~$+0.19$   &         & \tiny ~~~$+1.47$       &           & \tiny ~~~$+0.76$ \\
SP+KD           & 71.02     & 71.40                        & 74.98   & 75.63                             & 73.49     & 73.80      & 73.52     & 73.97            & 74.88   & 76.86                          & 68.41     & 69.37       \\[-0.4em]
                &           &\tiny ~~~${+0.38}$     &         &\tiny ~~~$+0.65$            &           &\tiny ~~~${+0.31}$ &           & \tiny ~~~$+0.45$    &         & \tiny ~~~$+1.98$    &           & \tiny ~~~$+0.96$ \\
CC+KD            & 70.88     & 71.31                        & 75.09   & 75.48                           & 73.04     & 73.46      & 73.48     & 70.79            & 74.71   & 76.29              & 68.02     & 68.39        \\[-0.4em]
             &           &\tiny ~~~${+0.43}$      &         &\tiny ~~~${+0.39}$        &           &\tiny ~~~${+0.42}$ &           & \tiny ~~~$+0.31$    &         & \tiny ~~~$+1.58$    &           & \tiny ~~~$+0.37$ \\
RKD+KD                  & 70.77     & 71.48                     & 74.89   & 75.61                            & 72.97     & 73.30      & 73.51     & 73.86    & 74.55   & 75.82       & 67.87     & 68.38        \\[-0.4em]
            &           &\tiny~~~$+0.71$         &         &\tiny ~~~${+0.52}$       &           &\tiny ~~~${+0.33}$ &           & \tiny ~~~$+0.35$    &         & \tiny ~~~$+1.27$             &           & \tiny ~~~$+0.51$ \\
VID+KD              & 71.10     & 71.53                  & 75.14   & 75.62                           & 73.19     & 73.47      & 73.46     & 73.88      & 74.85   & 75.97       & 68.27     & 68.75        \\[-0.4em]
             &           &\tiny ~~~${+0.43}$       &         &\tiny ~~~${+0.48}$        &           &\tiny ~~~${+0.28}$ &           & \tiny ~~~$+0.42$     &         & \tiny ~~~$+1.12$            &           & \tiny ~~~$+0.48$ \\
CRD+KD              & 71.56     & 71.95                        & 75.64   & 75.85                         & 74.29     & 74.60      & 74.58     & 74.98           & 76.05   & 76.93                    & 69.94     & 70.23        \\[-0.4em]
             &           &\tiny ~~~${+0.39}$       &         &\tiny ~~~${+0.21}$        &           &\tiny ~~~${+0.31}$ &           & \tiny ~~~$+0.40$   &         & \tiny ~~~$+0.88$       &           & \tiny ~~~$+0.29$ \\ 
REVIEW+KD        &~~71.78$^*$& 72.01                        &~~76.22$^*$   & 76.35                          &~~74.96$^*$     & 75.07       &~~73.89$^*$& 74.44       &~~77.89$^*$   & 78.10           &~~71.05$^*$     & 71.86       \\[-0.4em]
         &           &\tiny ~~~${+0.23}$   &         &\tiny ~~~${+0.13}$     &           &\tiny ~~~${+0.11}$ &           & \tiny ~~~$+0.55$   &         & \tiny ~~~$+0.21$      &           & \tiny ~~~$+0.81$ \\
        \bottomrule 
\end{tabular}}
\end{center}  
\vskip 0.1in
\end{table}

\noindent $\bullet$ \textbf{Results:}
The results are reported in \cref{table:cifar100}, where the upper-part of the table comprises the feature-based methods that are combined with KD, and the lower-part contains three logit-based KD variants.  

By noting the results in \cref{table:cifar100}, the following observations could be made:
\begin{itemize}
    \item Substituting the CE teacher with the MSE teacher in the benchmark KD methods consistently leads to an enhancement in the student's performance. This improvement can reach up to 2.67\%. 
    \item The improvement achieved in teacher-student pairs with different architectures is notably more substantial (these pairs are the last three columns of \cref{table:cifar100}).
    \item The accuracy of the teacher experiences a slight decline when employing the MSE loss. This observation underscores the distinction between training the teacher model for KD and training it solely for optimizing its individual performance. It affirms that these are distinct tasks, each with its own set of considerations and trade-offs.
\end{itemize}

\subsection{Additional Experiments on CIFAR-100} \label{sec:cifar2}
In this subsection, we follow the CRD paper \cite{CRD}, and combine the feature based distillation methods with conventional KD for achieving a higher performance. The results are listed in \cref{table:cifar100_2}. As seen, the MSE teacher consistently yield a better accuracy compared to its CE counterpart.

\begin{table}[ht] 
\begin{center}
\caption{Top-1 and Top-5 student's test accuracy (\%) on ImageNet validation set for 4 different KD methods using CE and MSE teachers (RN and MN stand for ResNet and MobileNet, respectively). \textbf{Bold} numbers and asterisk $(^*)$ to denote the best results and to identify the results reproduced on our local machines, respectively.} \label{table:imagenent}
\resizebox{1\textwidth}{!}{
\begin{tabular}{l|ccc|cc|cc|cc|cc}
\toprule
Teacher-Student           &         & \multicolumn{2}{c|}{Teacher Performance} & \multicolumn{2}{c|}{KD} & \multicolumn{2}{c|}{DKD} & \multicolumn{2}{c|}{REVIEW + KD} & \multicolumn{2}{c}{CRD +KD} \\ \hline
\multirow{3}{*}{RN34-R18}  &         & Top1         & Top5         & Top1       & Top5       & Top1                     & Top5  & Top1                  & Top5  & Top1       & Top5       \\ \hline
                          & CE     & 73.31        & 91.42        & 71.03      & 90.05      & 71.70                    &90.41 &~~71.84$^*$                 &~~90.77$^*$ & 71.38      & 90.49      \\
                          & MSE    & 71.66        & 90.83        & 71.58      & 90.77      & 71.93           & 91.23 & \textbf{72.16}                 & 90.93 & 71.57      & 90.36      \\ \hline
\multirow{2}{*}{RN50-MNV2} & CE     & 76.13        & 92.86       & 70.50      & 89.80      & 72.05                    & 91.05 & 72.56$^*$                 & 91.00$^*$ & 71.37$^*$      & 90.41$^*$      \\
                          & MSE    & 73.88        & 90.97       & 70.92      & 90.08      & 72.34                    & 91.17 & \textbf{72.91}        & 92.38 & 71.58      & 90.62 \\
                          \bottomrule
\end{tabular}} 
\end{center}
\vskip 0.1in
\end{table}

\subsection{ImageNet} \label{sec:imag}
ImageNet \cite{russakovsky2015imagenet} is a large-scale dataset used in visual
recognition tasks, containing around 1.2 million training and 50K validation images.

\noindent $\bullet$ \textbf{Teacher-student pairs:} 
Following the settings of \cite{CRD,yang2020knowledge}, we use 2 popular teacher-student pairs for our experiments (see \cref{table:imagenent}).

\noindent $\bullet$ \textbf{Training setup:}
We set the initial learning rate to 0.1 and divide the learning rate by 10 at 30, 60, and 90 epochs. We
follow the standard training process but train for 20 more
epochs (\ie, 120 epochs in total). Weight decay is set to 0.0001.

\noindent $\bullet$ \textbf{Results:}
We note that across all the knowledge transfer methods reported in \cref{table:imagenent}, replacing the CE teacher by MSE teacher consistently leads to an increase in the student accuracy. For example, when considering teacher-student pairs ResNet34-ResNet18 and ResNet50-MobileNetV2, the increase in the student's Top-1 accuracy in KD is 0.55\% and 0.42\%, respectively.

It is important to highlight that achieving such gains over the ImageNet dataset is considered substantial. Taken together, the results obtained across both CIFAR-100 and ImageNet datasets underscore the effectiveness of opting for MSE teachers over their CE counterparts.

\subsection{MSE teacher in semi-supervised distillation} \label{sec:semi}

\begin{wrapfigure}{r}{0.55\textwidth}
\vspace{-0.3in}
  \begin{center}
    \includegraphics[width=0.5\textwidth]{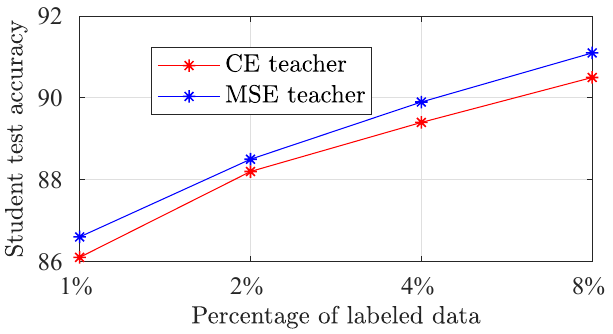}
  \end{center}
  \vspace{-5mm}
  \caption{The student's accuracy in semi supervised distillation for CE and MSE teachers.}
  \label{fig:semi}
  \vspace{-0.3in}
\end{wrapfigure}

Semi-supervised learning \cite{chen2020big,liu2022few, wu2023metagcd, zhong2022meta, iliopoulos2022weighted, chi2022metafscil,chi2021test,chi2020all} is a popular technique due to its ability to generate pseudo-labels for larger unlabeled dataset. In the context of KD, the teacher model has the responsibility of generating pseudo-labels for new, unlabeled examples (beyond its traditional function of providing soft targets during training).

To assess the MSE teacher's performance in a semi-supervised learning scenario, we conducted experiments using the CIFAR-10 dataset, following the settings outlined in \cite{chen2020big}, with the student model being ResNet18. In this experimental setup, although the dataset consists of 50,000 training images, only a small percentage of them are labeled—specifically, 1\% (500), 2\% (1000), 4\% (2000), and 8\% (4000). The outcomes are visualized in \cref{fig:semi}, where the student accuracy is plotted against the number of labeled samples. The results are averaged over three independent runs. As observed, the results demonstrate that not only can the MSE teacher effectively perform in a semi-supervised distillation scenario, but it also surpasses the performance of the CE teacher.

\subsection{Binary classification on customized CIFAR-$\{10,100\}$} \label{sec:binary}
The common understanding is that the enhancement in the accuracy of a student model in binary classification tends to be more modest compared to that observed in multi-class classification scenarios. This discrepancy arises due to the inherent limitations on the amount of information transferred from the teacher to the student network in binary classification settings, as documented in several studies \cite{sajedi2021efficiency, 9411995, tzelepi2021online, muller2020subclass, chiadapting, wu2024test}.

In this section, we want to empirically verify the effectiveness of MSE teacher in binary classification tasks. To this end, we create three binary classification datasets from CIFAR-$\{10,100\}$ datasets as explained in the sequel.

\noindent $\bullet$ \textbf{Dataset 1:} Following a similar approach to that described in \cite{muller2020subclass}, we construct the $\text{CIFAR}-2\times5$ dataset, wherein the input distribution exhibits a "sub-classes" structure. Specifically, we merge the first 5 classes of CIFAR-10 to form class one, while the remaining 5 classes constitute class two.

\noindent $\bullet$ \textbf{Dataset 2:}
We keep the training/testing samples from only two classes in the CIFAR-100 dataset: those belonging to class 20 and class 40, creating a dataset referred to as CIFAR-20-40.

\noindent $\bullet$ \textbf{Dataset 3:}
Similar to dataset 2, but we keep class 50 and 70  from CIFAR-100, which we refer to as CIFAR-50-70.

Now, we use VGG-13 and MobileNetv2-0.1 as the teacher and student models, respectively; and use conventional KD, AT, CC and VID to distill knowledge from CE and MSE teachers to the student.

The results for all three tasks are summarized in \cref{tab:binary}, with the reported values representing the average across five distinct runs. As seen, the MSE teacher yields a better student's accuracy all the cases. Also, it is worth noting that in some cases, for instance, AT over $\text{CIFAR}-2\times5$ dataset, the distillation method hurts the accuracy of the student.

\begin{table}[ht]
\caption{The student's accuracy (\%) in binary classification knowledge distillation on variants of CIFAR dataset. The results are averaged over five runs.}
\vspace{-.2in}
\label{tab:binary}
\begin{center}
\begin{tabular}{ccc|cc|cc}
\toprule
Dataset & \multicolumn{2}{c|}{$\text{CIFAR}-2\times5$} & \multicolumn{2}{c|}{CIFAR-26-45} & \multicolumn{2}{c}{CIFAR-50-74}  \\ \hline
Student Acc. & \multicolumn{2}{c|}{76.94}       & \multicolumn{2}{c|}{65.50}       & \multicolumn{2}{c}{66.30}        \\ \hline
Teacher    & CE      & MSE     & CE     & MSE     & CE     & MSE          \\ \hline
KD         & 77.00    & 77.34    & 69.70   & 70.15    & 71.40   & 71.23           \\
AT        & 67.39    & 68.21   & 64.70   & 65.33    & 67.10   & 69.12        \\
CC        & 77.19    & 77.64    & 70.30   & 70.83  & 69.70   & 71.38         \\
VID        & 77.16    & 77.47     & 69.25   & 69.88    & 69.80   & 69.97    \\
\bottomrule
\end{tabular}
\end{center}
\vskip -0.3in
\end{table}

\section{Conclusion}
This paper elucidated the significance of training the teacher model with MSE loss, which effectively minimizes the MSE between its output and BCPD. This approach aligns with the core responsibility of the teacher, namely, providing the student with a BCPD estimate that closely resembles it in terms of MSE. Through a comprehensive series of experiments, we demonstrated the efficacy of substituting the conventional teacher trained with CE loss with one trained using MSE loss in state-of-the-art KD methods. Notably, this substitution consistently enhanced the student's accuracy, leading to improvements of up to 2.6\%. In addition, we empirically showed the superior performance of MSE teacher in semi-supervised distillation task.

\clearpage
\bibliographystyle{splncs04}
\bibliography{egbib}
\end{document}